\pdfoutput=1
\documentclass{article}

    \PassOptionsToPackage{numbers, compress}{natbib}

\usepackage[final]{neurips_data_2024}

\usepackage[utf8]{inputenc} %
\usepackage[T1]{fontenc}    %
\usepackage{hyperref}       %
\usepackage{url}            %
\usepackage{booktabs}       %
\usepackage{amsfonts}       %
\usepackage{nicefrac}       %
\usepackage{microtype}      %
\usepackage{amsmath} 

\usepackage{multirow}
\usepackage[table,xcdraw]{xcolor}
\usepackage[normalem]{ulem}
\useunder{\uline}{\ul}{}

\usepackage{graphicx}
\usepackage{wrapfig}
\usepackage{tikz}
\usepackage{enumitem}
\usepackage{array}
\usepackage{multirow}
\usepackage{colortbl}
\usepackage{booktabs}
\usepackage{pythonhighlight}
\usepackage{listings}
\usepackage{tcolorbox}

\usepackage{verbatim}
\usepackage{listings}
\usepackage{xcolor}

\usepackage{wrapfig}
\usepackage{tikz}
\usepackage{enumitem}
\usepackage{array}
\usepackage{multirow}
\usepackage{colortbl}
\usepackage{booktabs}
\usepackage{pythonhighlight}
\usepackage{listings}
\usepackage{tcolorbox}
\usepackage{graphicx}

\lstdefinelanguage{json}{
  basicstyle=\ttfamily,
  numbers=left,
  numberstyle=\scriptsize,
  stepnumber=1,
  numbersep=8pt,
  showstringspaces=false,
  breaklines=true,
  frame=single,
  backgroundcolor=\color{white},
  literate=
    *{0}{{{\color{blue}0}}}{1}
     {1}{{{\color{blue}1}}}{1}
     {2}{{{\color{blue}2}}}{1}
     {3}{{{\color{blue}3}}}{1}
     {4}{{{\color{blue}4}}}{1}
     {5}{{{\color{blue}5}}}{1}
     {6}{{{\color{blue}6}}}{1}
     {7}{{{\color{blue}7}}}{1}
     {8}{{{\color{blue}8}}}{1}
     {9}{{{\color{blue}9}}}{1}
     {:}{{{\color{red}:}}}{1}
     {,}{{{\color{red},}}}{1}
     {\{}{{{\color{violet}{\{}}}}{1}
     {\}}{{{\color{violet}{\}}}}}{1}
     {[}{{{\color{violet}{[}}}}{1}
     {]}{{{\color{violet}{]}}}}{1},
}
\usepackage{graphicx}
\usepackage{tabularx}
\usepackage{arydshln}
\usepackage{tabularray}
\usepackage{tcolorbox}
\usepackage{geometry} %
\usepackage{adjustbox}
\usepackage{enumitem}

\definecolor{codegreen}{rgb}{0,0.6,0}
\definecolor{codegray}{rgb}{0.5,0.5,0.5}
\definecolor{codepurple}{rgb}{0.58,0,0.82}
\definecolor{backcolour}{rgb}{0.95,0.95,0.92}

\lstdefinestyle{mystyle}{
    backgroundcolor=\color{backcolour},  
    commentstyle=\color{codegreen},
    keywordstyle=\color{magenta},
    stringstyle=\color{codepurple},
    basicstyle=\ttfamily\footnotesize,
    breakatwhitespace=false,         
    breaklines=true,                 
    captionpos=b,                    
    keepspaces=true,                 
    showspaces=false,                
    showstringspaces=false,
    showtabs=false,                  
    tabsize=2
}

\definecolor{keycolor}{rgb}{0,0,0.8}    %
\definecolor{stringcolor}{rgb}{0.5,0,0} %
\definecolor{numbercolor}{rgb}{0.5,0,0.5} %

\definecolor{verylightgray}{rgb}{0.9,0.9,0.9}

\lstdefinelanguage{json}{
    basicstyle=\normalfont\ttfamily,
    showstringspaces=false,
    breaklines=true,
    frame=lines,
    backgroundcolor=\color{verylightgray},
    literate=
     *{0}{{{\color{numbercolor}0}}}{1}
      {1}{{{\color{numbercolor}1}}}{1}
      {2}{{{\color{numbercolor}2}}}{1}
      {3}{{{\color{numbercolor}3}}}{1}
      {4}{{{\color{numbercolor}4}}}{1}
      {5}{{{\color{numbercolor}5}}}{1}
      {6}{{{\color{numbercolor}6}}}{1}
      {7}{{{\color{numbercolor}7}}}{1}
      {8}{{{\color{numbercolor}8}}}{1}
      {9}{{{\color{numbercolor}9}}}{1}
      {:}{{{\color{keycolor}{:}}}}{1}
      {,}{{{\color{keycolor}{,}}}}{1}
      {\{}{{{\color{keycolor}{\{}}}}{1}
      {\}}{{{\color{keycolor}{\}}}}}{1}
      {[}{{{\color{keycolor}{[}}}}{1}
      {]}{{{\color{keycolor}{]}}}}{1}
      {"}{{{\color{stringcolor}{"}}}}{1},
}

\definecolor{lightgray}{rgb}{0.94,0.95,0.95}
\lstdefinelanguage{json2}{
    basicstyle=\normalfont\fontfamily{pcr}\selectfont,
    showstringspaces=false,
    breaklines=true,
    frame=lines,
    backgroundcolor=\color{lightgray},
    literate=
     *{0}{{{\color{numbercolor}0}}}{1}
      {1}{{{\color{numbercolor}1}}}{1}
      {2}{{{\color{numbercolor}2}}}{1}
      {3}{{{\color{numbercolor}3}}}{1}
      {4}{{{\color{numbercolor}4}}}{1}
      {5}{{{\color{numbercolor}5}}}{1}
      {6}{{{\color{numbercolor}6}}}{1}
      {7}{{{\color{numbercolor}7}}}{1}
      {8}{{{\color{numbercolor}8}}}{1}
      {9}{{{\color{numbercolor}9}}}{1}
      {:}{{{\color{keycolor}{:}}}}{1}
      {,}{{{\color{keycolor}{,}}}}{1}
      {\{}{{{\color{keycolor}{\{}}}}{1}
      {\}}{{{\color{keycolor}{\}}}}}{1}
      {[}{{{\color{keycolor}{[}}}}{1}
      {]}{{{\color{keycolor}{]}}}}{1}
      {"}{{{\color{stringcolor}{"}}}}{1},
}

\definecolor{mygray}{rgb}{0.95, 0.95, 0.95}
\definecolor{myblue}{rgb}{0.41, 0.50, 0.57}
\definecolor{greyblue}{RGB}{177,221,240}
\definecolor{lightgold}{RGB}{249,247,237}
\definecolor{peacockblue}{RGB}{27,161,226}

\tcbuselibrary{listings,skins,breakable}

\tcbset{
    enhanced,
    breakable,
    colback=mygray, %
    colframe=myblue, %
    boxrule=0.2mm, %
    arc=1mm, %
    top=10pt,
    bottom=10pt,
    left=10pt,
    right=10pt,
    listing only,
    listing options={
        basicstyle=\ttfamily\small, %
        language=Java, %
    },
}

\definecolor{light-gray}{gray}{0.95}

\title{xLAM: A Family of Large Action Models to \\ Empower AI Agent Systems}
\author{
\centerline{Jianguo Zhang\thanks{\ \   Co-first Authors.  },~ Tian Lan\footnotemark[1],~ Ming Zhu\footnotemark[1], ~\textbf{Zuxin Liu}\footnotemark[1], ~\textbf{Thai Hoang}\footnotemark[1], }\\ 
\centerline{~\textbf{Shirley Kokane}\thanks{\ \ Essential Contributors.}\footnotemark[2],~\textbf{Weiran Yao}\footnotemark[2],~\textbf{Juntao Tan},~\textbf{Akshara Prabhakar},~\textbf{Haolin Chen},~\textbf{Zhiwei Liu},} \\ \centerline{~\textbf{Yihao Feng},~\textbf{Tulika Awalgaonkar},~\textbf{Rithesh Murthy},~\textbf{Eric Hu},~\textbf{Zeyuan Chen},~\textbf{Ran Xu,}}\\ \centerline{~\textbf{Juan Carlos Niebles},~\textbf{Shelby Heinecke},~\textbf{Huan Wang\thanks{\ \ Corresponding Authors.}\footnotemark[3]},~\textbf{Silvio Savarese\footnotemark[3]},~\textbf{Caiming Xiong\footnotemark[3]}}\\\\
        \centerline{Salesforce AI Research} }

\begin{document}

\maketitle
\begin{figure}[h]
\vspace{-5mm}
\centering
\includegraphics[width=7cm]{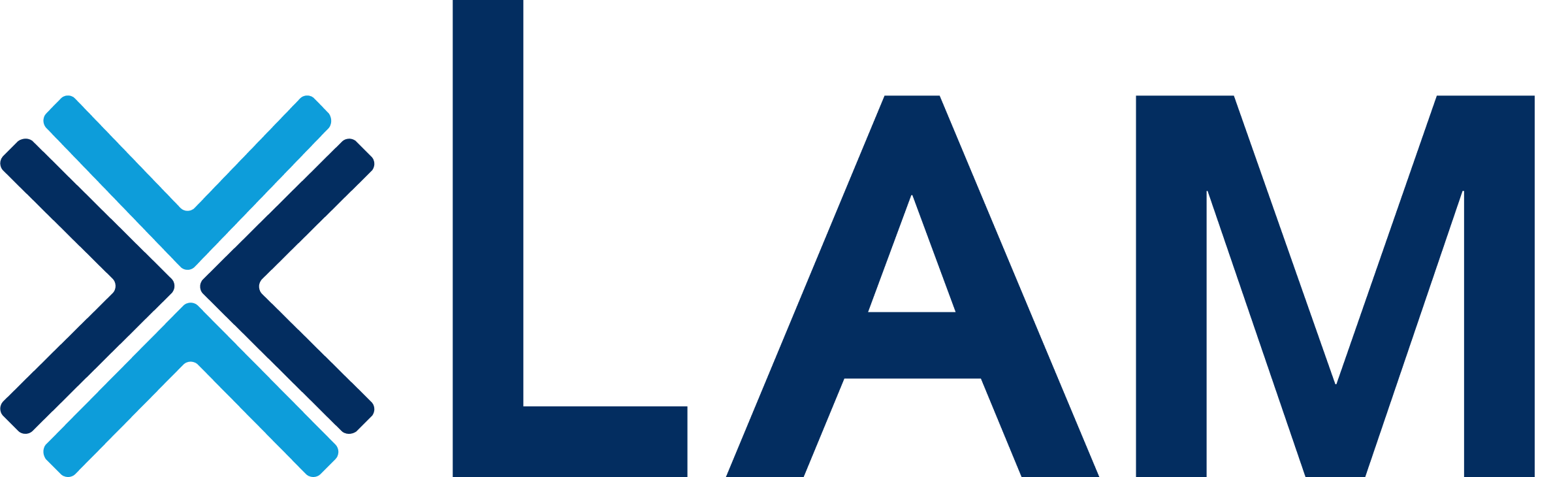}
\vspace{1mm}
\end{figure}

\begin{abstract}
Autonomous agents powered by large language models (LLMs) have attracted significant research interest. However, the open-source community faces many challenges in developing specialized models for agent tasks, driven by the scarcity of high-quality agent datasets and the absence of standard protocols in this area.
We introduce and publicly release \textbf{xLAM}, a series of large action models designed for AI agent tasks. The \textbf{xLAM} series includes five models with both dense and mixture-of-expert architectures, ranging from 1B to 8x22B parameters, trained using a scalable, flexible pipeline that unifies, augments, and synthesizes diverse datasets to enhance AI agents' generalizability and performance across varied environments.
Our experimental results demonstrate that \textbf{xLAM} consistently delivers exceptional performance across multiple agent ability benchmarks, notably securing the 1st position on the Berkeley Function-Calling Leaderboard, outperforming GPT-4, Claude-3, and many other models in terms of tool use. By releasing the \textbf{xLAM} series, we aim to advance the performance of open-source LLMs for autonomous AI agents, potentially accelerating progress and democratizing access to high-performance models for agent tasks.
\\\\
\textbf{Models:} \href{https://huggingface.co/collections/Salesforce/xlam-models-65f00e2a0a63bbcd1c2dade4}{huggingface.co/Salesforce/xLAM-models} \\
\textbf{GitHub:} \href{https://github.com/SalesforceAIResearch/xLAM}{github.com/SalesforceAIResearch/xLAM}\\

\end{abstract}

\section{Introduction}
The field of autonomous agents has witnessed significant advancements in recent years, with large language models (LLMs) playing a crucial role in enhancing agent capabilities across diverse tasks. Researchers have made substantial progress in developing sophisticated frameworks~\citep{hong2023metagpt,xagent2023,wu2023autogen,xie2023openagents} and specialized environments~\citep{deng2023mind2web,yao2022webshop,zhou2023webarena} to enhance agent capabilities, such as tool use~\citep{qin2023toolllm} and web browsing~\citep{zhou2023webarena}. Concurrently, comprehensive benchmarks like AgentBench~\citep{liu2023agentbench}, ToolBench~\citep{qin2023toolllm}, and AgentBoard~\citep{ma2024agentboard} have been established to rigorously assess agent performance in reasoning, planning, and multi-turn interactions.

While proprietary LLMs developed by industry leaders have demonstrated competitive performance in various agent tasks~\citep{anthropic2024claude,openai2023gpt4,reid2024gemini,touvron2023llama}, the open-source community faces limited choices for specialized models in this domain. This scarcity stems from several challenges in adapting open-source LLMs to agent tasks, primarily due to the lack of comprehensive, high-quality datasets and the heterogeneity of existing data formats. These factors complicate the unification of diverse datasets and obstruct the learning of transferable knowledge across different agent tasks.

Recently, the agent research community has intensified efforts in open-source agent data processing and model training~\citep{qin2023toolllm,chen2023fireact,xu2023lemur,patil2023gorilla,zeng2023agenttuning,yin2023lumos,zhang2024agentohana}. 
However, these works still face challenges in managing complex environments and generalizing to new scenarios, primarily due to limitations in the collected agent data.
A major obstacle is the homogeneity of content and format in existing datasets, resulting in models that lack diversity across various tasks and struggle to adapt to new or slightly different data structures in practical applications. While previous efforts have attempted to design pipelines for unifying data, they typically cover only a few scenarios or lack flexibility in their unified formats. For instance, Lumos~\citep{yin2023lumos} primarily addresses question answering, web agents, and mathematical tasks involving planning and grounding; while AgentOhana~\citep{zhang2024agentohana}, despite encompassing a more diverse range of environments, lacks an extendable unified format to accommodate new environments.

Moreover, open-source datasets often suffer from quality issues, such as incorrect agent outputs, hallucinated actions, and repeated interaction turns within trajectories~\citep{zhang2024agentohana, chen2024agentflan}. The lack of detailed analysis and understanding of agent data further complicates these challenges, hindering the development of robust and versatile open-source agent models. Addressing these challenges is crucial for advancing the field of open-source agent models and bridging the performance gap with proprietary LLMs in agent tasks.

\begin{figure*}[ht]
    \centering
    \includegraphics[width=\linewidth]{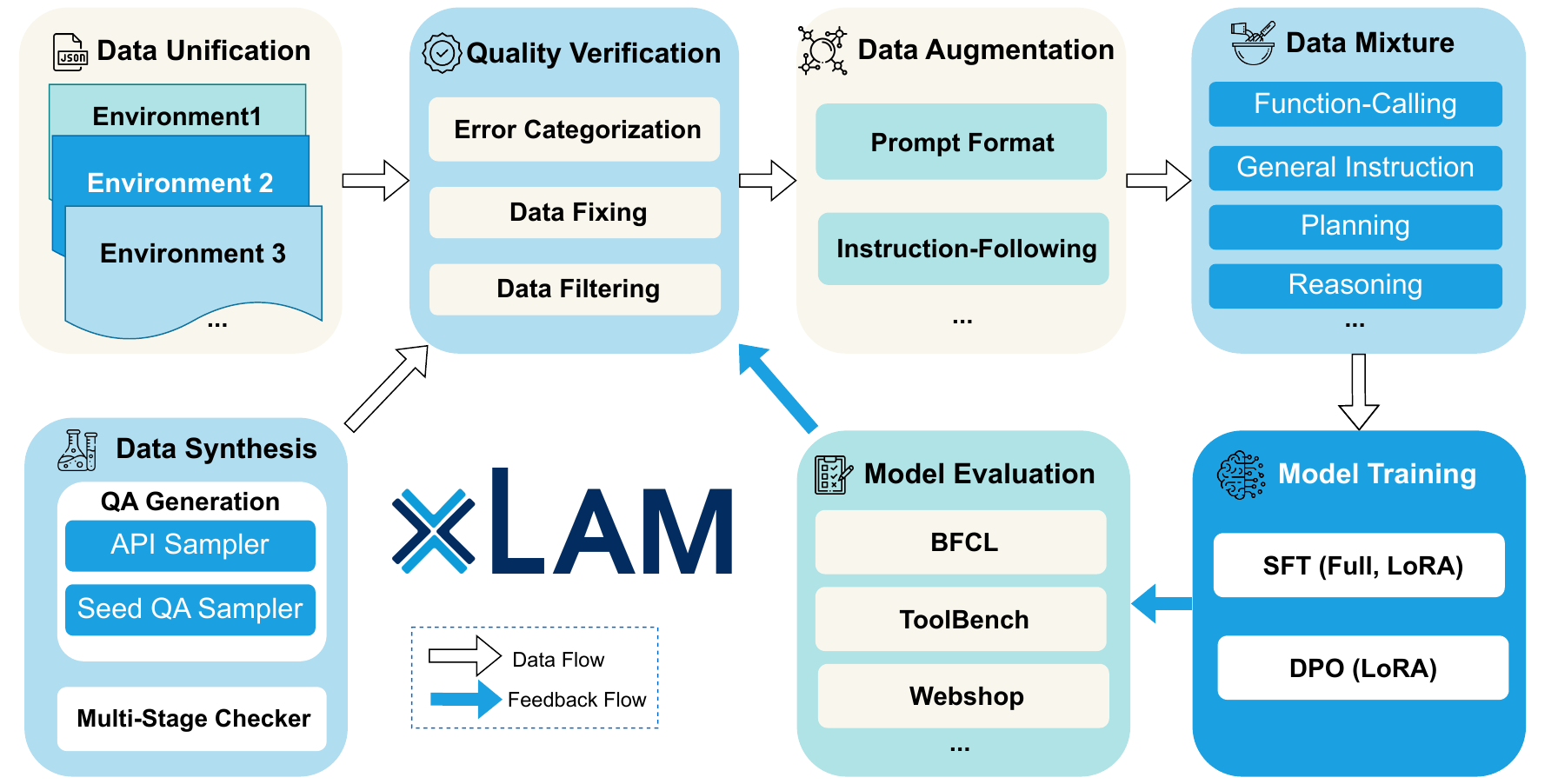}
    \caption{\small Overview of the data processing, training and evaluation of xLAM. We take the diagnostic feedback from the model evaluation results to iteratively improve the data quality.}
    \label{fig:data_pipeline}
\end{figure*}

\begin{figure}[ht]
    \centering
    \includegraphics[width=0.95\linewidth, trim=0cm 0cm 0cm 0cm, clip=true]{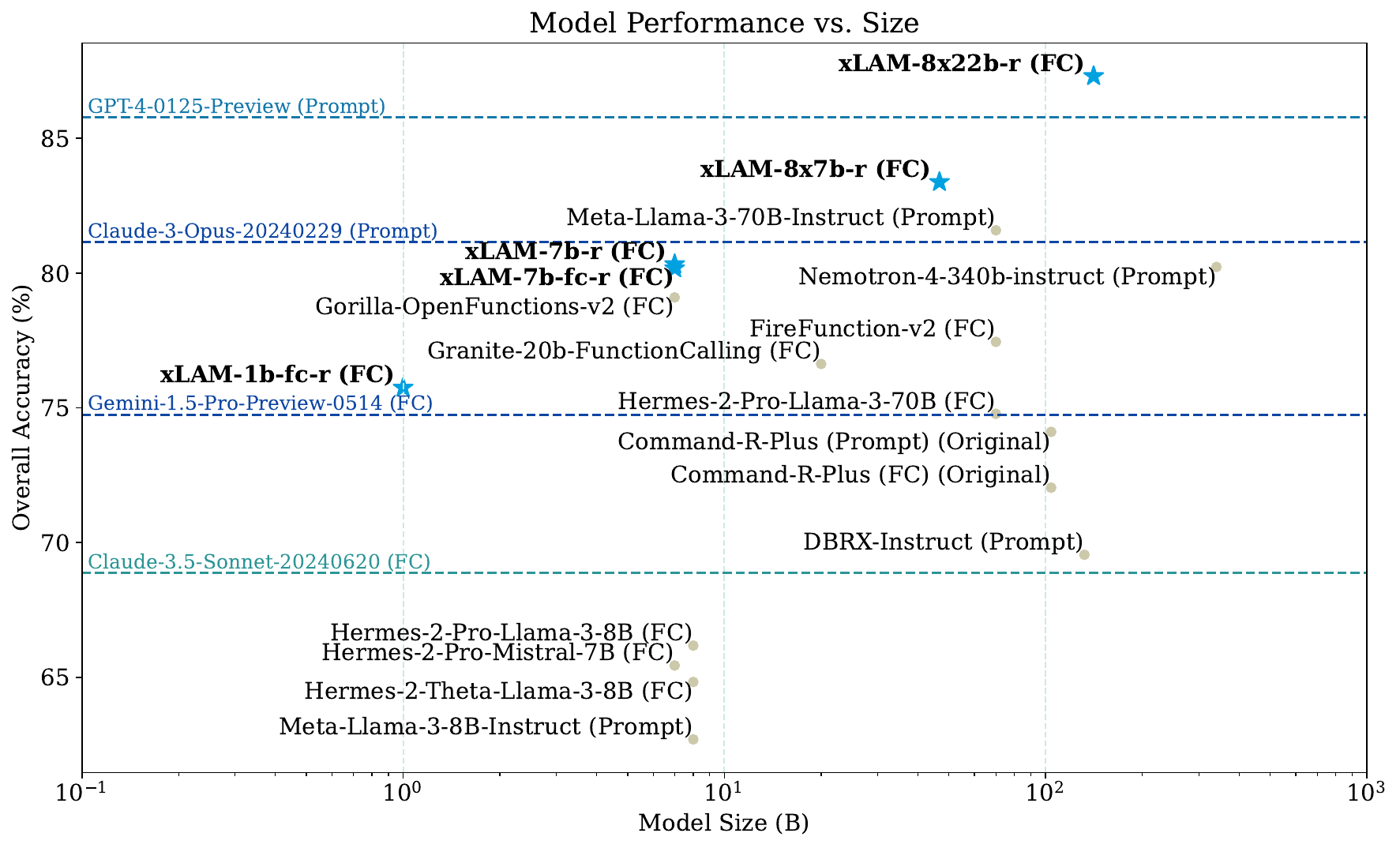}
    \caption{\small An overview of xLAM model performances on the Berkeley Function Calling Leaderboard v2 (cutoff date 09/03/2024). Our 8x22b model secures the top-1 position with wide margin on the leaderboard.}
    \label{fig:gorilla_rank}
\end{figure}

In this work, we introduce and open-source \textbf{xLAM}, a series of powerful models with varying sizes. This diverse set is tailored for a variety of applications, with smaller models (1B and 7B) optimized for on-device deployment, while larger models (8x7B and 8x22B) are designed to tackle more challenging tasks. Alongside the model release, we offer several insights and lessons learned from our experience in agent model training:

\begin{itemize}[leftmargin=*] 
    \item \textbf{Data Processing:} We highlight the importance of data unification and augmentation in enhancing dataset diversity and mitigating overfitting. Our developed dataset preprocess and augmentation pipeline significantly improves the generalizability of agent models across diverse environments.
    
    \item \textbf{Data Synthesis:} We showcase the impact of scalable, high-quality data synthesis on agent model performance. Our synthetic dataset enabled \textbf{xLAM} models to achieve 4 of the top 20 positions on the Berkeley Function Calling Leaderboard, including securing the top-1 spot (Fig. \ref{fig:gorilla_rank}), with smaller models achieving performance comparable to much larger counterparts, showing great potential in this direction.
\end{itemize}

We evaluate the \textbf{xLAM} series on public agent benchmarks, demonstrating exceptional performance across various agent tasks. By open-sourcing these models, we aim to advance open-source agent models and provide valuable insights into data processing and synthesis techniques, addressing key challenges in developing competitive alternatives to proprietary models.

\section{Related Work}

\subsection{LLM Agents}

Recent advancements in LLMs have significantly enhanced their utility in various agent tasks. Several innovative prompt techniques have been developed to improve performance, including Chain of Thought (COT)~\citep{wei2022chain}, ReACT~\citep{yao2023react}, and Reflection~\citep{shinn2023reflexion}. Additionally, considerable efforts have been made to fine-tune open-sourced agent models for better capabilities~\citep{qin2023toolllm, chen2023fireact,patil2023gorilla,  zeng2023agenttuning, zhang2024agentohana}. These include enhancements in data collection and processing to facilitate effective agent learning~\citep{zeng2023agenttuning, li2023api, tang2023toolalpaca, yin2023lumos, zhang2024agentohana, chen2024agentflan}, covering a range from simple question answering to more complex scenarios like web interactions, tool operations, reasoning, and planning. 
However, many of these agent frameworks still depend on proprietary models as their core engine to achieve optimal performance, revealing a substantial gap in the availability of high-quality open-source models for these tasks.

\subsection{Agent Benchmarks}

A variety of benchmarks have been established to assess the abilities of LLM agents across diverse scenarios~\citep{yao2022webshop,  qin2023toolllm, liu2023agentbench,
ma2024agentboard,huang2023metatool, liu2023bolaa, wang2023mint, liu2024agentlite, du2024anytool,  bfcl}. Notably, AgentBench~\citep{liu2023agentbench}, Mint-Bench~\citep{wang2023mint}, and AgentBoard~\citep{ma2024agentboard} encompass environments ranging from code generation and games to web interactions and reasoning tasks. ToolBench~\citep{qin2023toolllm} specifically evaluates multi-turn reasoning and tool-usage abilities, while the Berkeley Function-Calling Leaderboard~\cite{bfcl} broadly assesses models' capabilities in function calling across various contexts. These recent advancements in benchmarking have made the evaluation of agent models more accessible and standardized.

\section{Data Processing Pipeline}
\label{sec:data}

In this section, we discuss the data pipeline for training xLAM, including data unification, augmentation, quality verification, general instruction data synthesis, and preference data generation.

\subsection{Data Unification}

Existing agent datasets are collected from diverse environments and designed in various formats, introducing noise and complicating data augmentation and verification. Models like NexusRaven~\cite{srinivasan2023nexusraven}, Gorilla-Openfunctions~\cite{gorilla-openfunctions-v2}, and AgentOhana~\cite{zhang2024agentohana} have demonstrated superior performance in function-calling, suggesting that a well-defined, universal format could significantly enhance model performance. By standardizing the format of existing data, we can reduce noise and facilitate easier data augmentation and quality verification, leading to a more efficient and robust framework for model training and evaluation. Furthermore, a standardized format ensures consistency, simplifies model training, and enhances the model's ability to generalize across various benchmarks.

Function-calling formats form the basis for how models understand and execute tasks, motivating us to design our unified data format in a function-calling style. 
As illustrated in Figure \ref{fig:unified_format}, 
the unified format consists of several modules: task instruction, available tools, format instruction, few-shot examples, query, and steps. 
Specifically, the available tools define the agent's action space, and the format instruction specifies the output format the agent should follow when generating a response. In each step, the agent's output, the environment's feedback/execution results, and the user's follow-up input are organized into a dictionary. 
It's quite common for there to be purely conversational interactions between users and agents that don't trigger any APIs or receive corresponding observations. In these instances, the related entry values would simply remain empty.

This unified format is compatible with various environments and tasks, making our data processing pipeline adaptable to different datasets and scalable to large amounts of data. Moreover, the modularized design allows for fine-grained data augmentation and quality verification, which are essential in improving agent data quality. For example, by unifying all the available tools and tool calls, we can easily inspect for hallucination and function-call errors, and apply various augmentation techniques.

\subsection{Data Augmentation}
\label{ss:data_augmentation}

Our data augmentation strategy focuses on improving the diversity of the data. It involves applying various transformations to the existing dataset, thereby generating new, synthetic data samples. The data unification step significantly simplifies the application of various augmentation techniques. A standardized data format ensures consistency and ease of implementation, allowing for more efficient augmentation processes. Specifically, the augmentation techniques we adopted can be categorized as prompt format augmentation and instruction-following augmentation.

\textbf{Prompt Format Augmentation:} Prompt format augmentation focuses on creating various prompt formats based on the structured, unified data format. The format augmentation can be further divided into two categories: 1) \textit{Order Shuffling}. In the unified format, the available tools are provided in a list, and each tool contains the name, description, and parameters. To avoid model overfitting to the specific order of the tools, we randomly shuffle the tool list. Furthermore, we also shuffle the order of the name, description, parameters, and within the parameters to present the information in different ways. We do the same thing within the tool\_calls in each step. Additionally, we also shuffle the order of different sections of the input, including task instruction, tools, format instruction, few-shot examples etc. 2) \textit{Concatenation Tokens}. Each training data point is a pair of input and output sequences. To convert the structured unified format to the training prompt, we use special tokens to concatenate different sections into one sequence. We create several different special token styles, including "[START/END OF QUERY]", "<query></query>", and plain text. 

\textbf{Instruction-Following Augmentation:}
Instruction-following augmentation focuses on adding diversity to the instructions in order to improve the model's instruction-following capability. It involves rephrasing existing instructions and adding new instructions, without introducing inaccuracy and inconsistency. Therefore, verification of the new instructions is a crucial step for this type of augmentation. We employ two methods for instruction-following augmentation: 1) \textit{Task Instruction Rephrasing.} We rephrase the task instructions using powerful LLMs to accommodate various input styles from users. To ensure the rephrased instructions still align with the original version, we verify them by prompting the LLMs with the rephrased instructions and check if the LLMs can still follow them and generate correct function calls. 2) \textit{Format Instruction-Following.} In our unified format, the output format is a JSON string with \texttt{thought} and \texttt{tool\_calls}. To avoid the model overfitting on JSON format and to enable the model to follow various output formats upon different format instructions, we prepare 15 different output formats along with their corresponding format instructions and format converters. The output formats include JSON, XML, YAML, plain text, etc.

\subsection{Data Quality Verification}
\label{sec:data-verification}
To further understand of the data quality and to thoroughly investigate the sources of errors in the evaluation, we conduct a detailed analysis of the unified dataset. We identify a list of errors in the data using both rule-based and LLM-as-a-judge approaches.

\textbf{Undefined Function Call:} 
In function-calling, a list of available functions is provided, and the model should generate a function\_call using one of the given functions. However, we found that in many cases, the predicted function\_call is not from the given list. We match the predicted function with the given functions by comparing the function names and the list of parameter names. When the function\_call name does not match any given functions, we refer to it as \textit{Undefined Functions Invoked}. When the function name matches but the argument list contains undefined arguments, we refer to it as \textit{Undefined Arguments Passed}. We also take into consideration optional parameters.

\textbf{Incorrect Argument Type:}
Other than the error types mentioned above, we also observe that sometimes the model generates the correct argument's value, but in the wrong types. For example, when a parameter expects a \texttt{[val1, val2, val3]}, the generated arguments is \texttt{"[val1, val2, val3]"}, which is a string version of the list. When executing the function call, errors will occur due to incorrect data type.  We identify trajectories containing the incorrect argument type error by comparing the parameter type in the available tools and the actual argument type. We also found that most argument type errors can be fixed by converting the arguments to the correct parameter types. 

\textbf{Argument Hallucination:} Upon examining the unified dataset from public sources, we discovered that tool calls frequently include argument values not present in the user query or prior steps. This issue arises because much of this data is generated by LLMs, which are prone to hallucination, a common problem in LLM-generated content.
We identified two types of hallucination: 1) the generated tool names or argument names do not appear in the provided tool and argument list; and 2) the argument values do not align with the user query or observations from previous steps. The first type of hallucination is straightforward to address by searching the generated tool call and argument names and matching them with the provided tool list, as they are all structured in JSON, making this process efficient. However, detecting the second type, where argument values are misaligned, is more challenging, as simple string matching is ineffective for complex queries and tasks. To tackle this, we use LLMs as judges to perform step-wise argument hallucination detection, detecting if there is a mismatch between the arguments and the intended query or prior observations.

\textbf{Low-Quality Reasoning and Planning:} We observe many data trajectories where the reasoning and planning steps are of low quality, which is a common issue in the outputs of many LLMs. To address this, we first filter out low-quality data using rule-based methods informed by heuristics, then prompt models like Mixtral-8x22b-Instruct-v0.1~\citep{jiang2024mixtral} and DeepSeek-V2~\citep{deepseekv2} to evaluate both the overall trajectory and individual thought steps on the selected data. A portion of these rating results is then sampled and verified by humans. We also attempted to iterate on this process using specifically fine-tuned models.

\subsection{Data Synthesis}

Based on our findings in Sec. \ref{sec:data-verification}, we observe that most of these publicly available datasets have several limitations. First, these datasets are often static, synthesized by weak models, limited in scope, and, more importantly, not verified by execution. 
Second, these datasets mainly focus on a single type of function-calling category, i.e., outputting a single function call based on the provided tools. However, real-world scenarios might consist of many other types of use cases, such as the parallel function-calling scenario \cite{bfcl}, where the user query contains multiple requests and the model should respond with concurrent function calls in parallel within a single response.

To address these two issues, we utilize a systematic data synthesis framework called APIGen \cite{liu2024apigen}, which can generate verifiable datasets based on a collection of executable APIs.
The key idea is a multi-stage verification process to ensure the accuracy and quality of the generated data. This process includes format verification as introduced in Sec. \ref{sec:data-verification}, execution verification, and semantic verification, which collectively help to identify and filter out low-quality data points, such as those with hallucination issues or inaccurate argument parameter values.

We utilize over 3,673 APIs across 21 categories from ToolBench \cite{qin2023toolllm} to generate a total of 60,000 high-quality data. These samples are generated using several strong open-source language models: DeepSeek-V2-Chat \cite{deepseekv2} and Mixtral-8x22B-Inst \cite{jiang2024mixtral}. 
This synthesis framework greatly improves the robustness and applicability of the dataset, as the majority of low-quality data can be identified by the multi-stage verification process.

\subsection{Data Mixture}
For supervised fine-tuning (SFT), our dataset combines training samples from three main sources: cleaned and augmented agent datasets, a synthetic function-calling dataset, and general instruction-tuning datasets. These sources are used to train the general xLAM models.  

Specifically, to enhance the general instruction capability of xLAM, we integrate diverse instruction-tuning datasets from DialogStudio~\citep{zhang2023dialogstudio} and Data Provenance~\citep{longpre2023data,longpre2024consent}. We employe rule-based techniques to filter out low-quality data, such as repetitive words and turns, which are common and often produced by less powerful models. We also remove data with inappropriate contents, responses and non-commercial licenses.  Additionally, we deduplicate examples with similar user queries and organized the data by domain or category. We then prompt  Mixtral-8x22b-Instruct-v0.1 and DeepSeek-V2 to assess both the entire dialogue and individual system responses on the selected data. This instruction data comprises 20\% to 30\% of our training set.  To further enhance model robustness, we preserve the original formats of the general instruction-tuning data. 

To enhance the function-calling capability of xLAM-7b-fc-r and xLAM-1b-fc-r, we employ a targeted training approach, with 50\% of their training data drawn from our high-quality synthetic function-calling dataset. The remaining 50\% of the training data is sampled from other tasks within our training set.

For Direct Preference Optimization (DPO)~\citep{rafailov2023direct}, we prompt less powerful models to generate and rate responses for selected data from each source, then sample a subset for human verification. After adjustments to models and prompts, we classify the selected responses as rejected samples.

\section{Model Training}

\subsection{Modeling}\label{sec:training-data}

We use a supervised fine-tuning 
(SFT) approach, further aligning model checkpoints with the DPO method, and leverage the robustness of our flexible data pipeline. Our training code is based on the HuggingFace Transformers and Accelerate libraries\citep{wolf2020transformers, accelerate}, as well as PyTorch FSDP\citep{zhao2023pytorchfsdpexperiencesscaling}. During training, the model undergoes multiple epochs, with datasets randomly shuffled each time. When using data parallelism across multiple devices, we diversify random seeds based on process IDs, ensuring balanced data distribution through partitioning, shuffling, and interleaving, thereby enhancing the robustness and reproducibility of our training process.

The fine-tuning of general xLAM models is conducted on Nvidia H100 GPUs. For SFT, we use a full fine-tuning framework that employs the fully sharded data parallel algorithm \citep{zhao2023pytorch}. In the case of xLAM-8x22b-r, we integrate LoRA \citep{hu2021LoRA,dettmers2023qLoRA} to better preserve the model's original capacities and prevent catastrophic forgetting \citep{liu2023tail}. LoRA is also used for DPO alignment across all xLAM models. Additionally, we use a cosine learning rate scheduler with 100 warm-up steps to optimize performance.

The xLAM-FC models target various categories of function-calling agents, including simple, multiple, parallel, and parallel multiple. These categories are designed to enhance the models' performance in different scenarios. For instance, a simple query like retrieving the weather for a location (e.g., "What is the weather in Palo Alto today?") can be handled by calling \texttt{get\_weather("Palo Alto", "today")}. Multiple queries involve selecting the appropriate function from several APIs, while parallel queries require executing multiple function calls simultaneously. Additionally, the models are trained in relevance detection to ensure alignment between function calls, execution results, and query objectives.

\begin{table*}[ht]
\centering
\small
\resizebox{0.95\textwidth}{!}{%
\begin{tabular}{lcccc}
\toprule
Model & Base Model & \# Total Params & Context Length & Category \\
\midrule
xLAM-1b-fc-r      & DeepSeek-Coder-1b   & 1.35B           & 16k            & Function-calling \\
xLAM-7b-fc-r      & DeepSeek-Coder-7b   & 6.91B           & 4k             & Function-calling \\
xLAM-7b-r         & Mistral-7b          & 7.24B           & 32k            & General          \\
xLAM-8x7b-r       & Mistral-8x7b        & 46.7B           & 32k            & General          \\
xLAM-8x22b-r      & Mistral-8x22b       & 141B            & 64k            & General          \\
\bottomrule
\end{tabular}%
}
\caption{\small Overview of xLAM model series.}
\label{tab:xlam_series}
\end{table*}

\subsection{xLAM Model Series}
We introduce a series of agent models tailored for different use cases. Our flagship model series, xLAM, is built upon the Mixtral Instruct \cite{jiang2024mixtral} models and aims to achieve balanced performance across a diverse range of agent tasks, from complex multi-turn interactions to function-calling applications. To ensure its versatility, xLAM is trained on uniformly sampled data from our training dataset as introduced in Sec. \ref{sec:training-data}.

In addition to general xLAM models, we develop two specialized models for function-calling use cases, xLAM-7b-fc-r and xLAM-1b-fc-r, based on DeepSeek-Coder-7B-instruct-v1.5 and DeepSeek-Coder-1.3B-instruct, respectively \cite{guo2024deepseek}. The smaller model sizes offer increased accessibility, allowing users to easily host them on a single GPU to address various function-calling tasks, ranging from simple user queries to parallel concurrent requests.

By offering a suite of models with varying sizes and specializations, the xLAM series caters to a wide range of user needs and computational resources, making powerful agent capabilities more accessible and adaptable to real-world applications.

\section{Experiments}

\subsection{Benchmarks}

After considering the stability of environments and research budget limitations, we evaluate the performance of models across four rigorous benchmarks: Webshop \citep{yao2022webshop}, ToolQuery \citep{ma2024agentboard}, ToolBench \citep{qin2023toolllm}, and the Berkeley Function-Calling Benchmark \citep{bfcl}. Each benchmark is designed to assess different aspects of model capabilities under a variety of settings and constraints.

\noindent \textbf{Webshop} is an interactive web environment designed to mimic online shopping experiences, testing an agent's ability to navigate and assist in e-commerce tasks. Webshop comprising approximately 250 test cases. 

\noindent \textbf{ToolQuery} evaluates an agent's skills in using tools to retrieve and process information across domains. ToolQuery features 60 test cases across three distinct settings: Weather, Movie, and Academia.

We use the testing configurations from AgentBoard \citep{ma2024agentboard} for both Webshop and ToolQuery. These configurations assess overall performance using the Success Rate and evaluate progressive performance across interactive turns with the Progress Rate, with  Success Rate being the more critic metric.

We additionally evaluate on \textbf{ToolQuery-Unified}, which is essentially ToolQuery but requires an agent to ingest the task instruction and tools following the augmented prompt format described in \S\ref{ss:data_augmentation}  and likewise solve the task following the unified format. 
The purpose of testing agents in this setting is to assess their reasoning and tool-use abilities when evaluated on structured formats \citep{tam2024let}.

\noindent \textbf{ToolBench} is developed for real-time evaluation of multi-turn reasoning and interactive capabilities via RapidAPI, and includes around 1,000 test cases. It uses Pass Rate as the metric, where the trajectory and final response are sent to GPT-4-0125-preview to determine whether the agent's final response successfully addresses the given user query. The evaluations cover both in-domain and out-of-domain settings, including unseen instructions with familiar tools, unseen tools within previously known categories, and entirely new categories of unseen tools.

\noindent \textbf{Berkeley Function-Calling Leaderboard (BFCL) Benchmark} \cite{bfcl} provides a comprehensive evaluation framework for assessing an agent's capability to reason about and execute function calls across a variety of programming languages and application domains. The benchmark comprises over 2,200 test cases, challenging models with complex scenarios such as parallel and multiple function calls in languages like Java, JavaScript, and Python. The evaluation metrics include Abstract Syntax Tree (AST) accuracy for non-executable test queries, executable accuracy by running APIs to obtain results, and a relevance detection score that measures the agent's ability to distinguish non-relevant queries and provided tools.

\noindent Importantly, our evaluation utilizes the most recent BFCL v2 version, as of the cutoff date 09/03/2024. The v2 version introduces live function calls and real-world scenarios contributed by users, addressing issues such as data contamination, bias, and fairness by leveraging user-provided data. This updated dataset better reflects real-world distributions, characterized by a higher demand for selecting among multiple functions and a reduced demand for parallel function calls. For instance, our analysis indicates that in the v2 benchmark, the average number of available functions has doubled, while the average number of function calls has been halved compared to the non-live v1 data. It is important to note that all our models were trained prior to the release of the BFCL v2 live data.

\subsection{Experimental Results}

\subsubsection{Webshop and ToolQuery}

\begin{table*}[ht]
\centering
\small
\resizebox{0.9\textwidth}{!}{%
\begin{tabular}{@{}lcccccc@{}}
\toprule
\multirow{2}{*}{LLM} & \multicolumn{2}{c}{Webshop} & \multicolumn{2}{c}{ToolQuery}  \\ \cmidrule(l){2-5} 
                       & Success Rate       & Progress Rate    & Success Rate       & Progress Rate          \\ \midrule
\textbf{xLAM-7b-r } & \textbf{0.414} & 0.767 & 0.550 & 0.674 \\
\textbf{xLAM-8x7b-r } & \underline{0.410} & 0.763 & \underline{0.683} & 0.745 \\
\textbf{xLAM-8x22b-r } & 0.390 & 0.763 & \underline{0.683} & 0.758 \\
\midrule
GPT-4-0125-preview     &  0.375 & 0.760 & \textbf{0.750} & 0.803 \\
GPT-4o-2024-0523 & 0.323         & 0.694         & 0.633          & 0.801             \\
AgentOhana-8x7b~\citep{zhang2024agentohana} & 0.331 & 0.737 & 0.533 & 0.766 \\
Claude2 & 0.378 & 0.746 &  0.483 & 0.735 \\
Mixtral-8x22b-inst~\citep{jiang2024mixtral} & 0.383 & 0.739 & 0.400  & 0.740 \\
DeepSeek-67b-chat~\citep{bi2024deepseek} & 0.319 & 0.727 & 0.400 & 0.714 \\
GPT-3.5-Turbo-0125 & 0.323    & 0.749         & 0.367          & 0.545             \\
Lemur-70b-chat-v1~\citep{xu2023lemur} & 0.116 & 0.718 & 0.283 & 0.720 \\
Mixtral-8x7b-inst~\citep{jiang2024mixtral} & 0.222 & 0.766 & 0.167  & 0.654 \\ 
CodeLlama-34b-inst~\citep{roziere2023code} &  0.235 & 0.717 & 0.133 & 0.600 \\
Llama2-70b-chat~\citep{touvron2023llama} & 0.131 & 0.536 &  0.000 & 0.483 \\
Vicuna-13b-16k~\citep{chiang2023vicuna} & 0.219 & 0.733 & 0.033 & 0.343 \\ 
\bottomrule
\end{tabular}%
}\caption{\small Testing results on Webshop and ToolQuery. \textbf{Bold} and \underline{Underline} results denote the best result and the second best result for Success Rate, respectively. }\label{tab:agentboard_results}
\end{table*}

\textbf{Webshop.} Table \ref{tab:agentboard_results} presents detailed comparisons of state-of-the-art language and agent models in the Webshop and ToolQuery environments, illustrating the robust and strong performance of the xLAM models. In the Webshop environment, xLAM-7b-r not only achieves the highest Success Rate at 0.414, surpassing other general LLMs like GPT-4-0125-preview, GPT-4o-2024-0523, and Claude2, but also outperforms specialized agent models such as AgentOhana-8x7b and Lemur-70b. This demonstrates xLAM models' superior ability to navigate and execute tasks in the web interaction environment effectively.

\textbf{ToolQuery.} In the more complex and unseen ToolQuery environment, xLAM-8x7b-r and xLAM-8x22b-r also demonstrate high performance as shown in Table \ref{tab:agentboard_results}, ranking second with a Success Rate of 0.683. This shows a significant improvement over the baseline performance of Mixtral-8x7b-inst and Mixtral-8x22b-inst, which are 0.167 and 0.400, respectively. Notably, all three xLAM models surpass the Mixtral-8x22B-Instruct model. Despite Mixtral-8x22B-Instruct having a large number of parameters and specialized tuning for advanced functionalities such as function calling, reasoning, and complex tool usage, it falls short of the xLAM models’ performance. Furthermore, same as other general LLMs, it lacks transparency regarding the data collection, unification processes, and other critical details, contrasting with the open source purposes provided for xLAM. These results show the efficacy of our proposed data unification and synthetic data pipeline. 

\begin{table*}[ht]
\centering
\small
\resizebox{0.7\textwidth}{!}{%
\begin{tabular}{lcccc}
\toprule
& Success Rate & Academia & Movie & Weather \\
\midrule
\textbf{xLAM-7b-r}          & 0.466 (0.550)        & 0.45     & 0.25  & 0.35    \\
\textbf{xLAM-8x7b-r}        & 0.533 (0.683)        & 0.45     & 0.40   & 0.45    \\
\textbf{xLAM-8x22b-r}       & \textbf{0.733} (0.683) & 0.75   & 0.40   & 0.60     \\
\midrule
GPT-4-0125-preview & \underline{0.566} (0.750) & 0.65  & 0.35  & 0.25    \\
GPT-4o-2024-05-13  & 0.366 (0.633)        & 0.45     & 0.20   & 0.25    \\
\bottomrule
\end{tabular}%
}
\caption{\small Testing results on ToolQuery-Unified. \textbf{Bold} and \underline{Underline} results denote the best result and the second best result for Success Rate, respectively. Values in brackets indicate corresponding performance on ToolQuery.}
\label{tab:toolquery_unified_results}
\end{table*}

\textbf{ToolQuery-Unified.} When the system prompt from ToolQuery is presented to the model in the unified format shown in Fig. \ref{fig:training_template}, and the model is required to follow the provided format instructions to generate a structured output, we observe that xLAM models' performances are more consistent compared to GPT models, as shown in Table \ref{tab:toolquery_unified_results}. While GPT-4o’s performance significantly degrades by 42\% compared to ToolQuery, our best xLAM 8x22b model maintains comparable performance. This can be attributed to xLAM being trained on trajectories that adhere to the unified format, enabling it to perform consistently during inference. Concurrent research \cite{tam2024let} observed a similar decline in performance on reasoning tasks when LLMs are constrained to produce output in specific formats. Deeper analysis indicated that the degradation is more than just due to incorrectly formatted output in a specific format, but rather due to a drop in the reasoning ability of the model itself.

\begin{table*} %
\resizebox{1.0\linewidth}{!}{
\centering
\begin{tabular}{lcccc}
\toprule
    & \begin{tabular}[c]{@{}c@{}}Unseen Insts \& Same Set \end{tabular} &  \begin{tabular}[c]{@{}c@{}} Unseen Tools \& Seen Cat\end{tabular} & \begin{tabular}[c]{@{}c@{}}Unseen Tools \& Unseen Cat \end{tabular}  \\ 

\midrule
\textbf{xLAM-7b-r } & \underline{0.5308} & \underline{0.5300} & \textbf{0.5850}	 \\
\textbf{xLAM-8x7b-r } & \underline{0.5308} & \textbf{0.5450} &	\underline{0.5700}	\\
\textbf{xLAM-8x22b-r } & - & - & - \\
\midrule
AgentOhana-8x7b~\citep{zhang2024agentohana}            & 0.5077  & 0.5200 & 0.5650                  \\ 
GPT-4-0125-preview              & \textbf{0.5462} & 0.5050   & 0.5450           \\ 

GPT-3.5-Turbo-0125  & 0.5000 & 0.4900   & 0.5150                     \\

TooLlama-V2~\citep{qin2023toolllm}      & 0.4385  & 0.4350  & 0.4300                      \\

\bottomrule
\end{tabular}
}
\caption{\small Pass Rate on ToolBench on three distinct scenarios. 
\textbf{Bold} and \underline{Underline} results denote the best result and the second best result for each setting, respectively. The results for xLAM-8x22b-r are unavailable due to the ToolBench server being down between 07/28/2024 and our evaluation cutoff date 09/03/2024.
}
\label{table:tool-eval}
\end{table*}

\subsubsection{ToolBench}

Table \ref{table:tool-eval} presents the results on ToolBench, where xLAM models demonstrate impressive performance. They surpass both TooLlama-V2 and GPT-3.5-Turbo-0125 across all test settings. Moreover, xLAM models outperform AgentOhana-8x7b in scenarios involving unseen instructions and unseen tools, while achieving performance comparable to GPT-4-0125-preview in the unseen tools setting. These results show xLAM models' robust capabilities in multi-turn reasoning and complex tool usage, effectively handling both in-domain and out-of-domain tasks.

\subsubsection{Berkeley Function-Calling Benchmark}

\begin{table}[h]
\centering
\resizebox{\linewidth}{!}{
\begin{tabular}{|c|c|c|cccc|cccc|cc|}
\hline
\multirow{4}{*}{Rank} & \multirow{4}{*}{\begin{tabular}[c]{@{}c@{}}Overall\\ Accuracy\end{tabular}} & \multirow{4}{*}{Model}             & \multicolumn{4}{c|}{\multirow{2}{*}{Abstract Syntax Tree (AST) Evaluation}}                                                                                    & \multicolumn{4}{c|}{\multirow{2}{*}{Evaluation by Executing APIs}}                                                                                             & \multicolumn{2}{c|}{\multirow{2}{*}{\begin{tabular}[c]{@{}c@{}}Relevance\\ Detection\end{tabular}}} \\
                      &                                                                             &                                    & \multicolumn{4}{c|}{}                                                                                                                                          & \multicolumn{4}{c|}{}                                                                                                                                          & \multicolumn{2}{c|}{}                                                                               \\ \cline{4-13} 
                      &                                                                             &                                    & \multirow{2}{*}{Simple} & \multirow{2}{*}{Multiple} & \multirow{2}{*}{Parallel} & \multirow{2}{*}{\begin{tabular}[c]{@{}c@{}}Parallel\\ Multiple\end{tabular}} & \multirow{2}{*}{Simple} & \multirow{2}{*}{Multiple} & \multirow{2}{*}{Parallel} & \multirow{2}{*}{\begin{tabular}[c]{@{}c@{}}Parallel\\ Multiple\end{tabular}} & \multirow{2}{*}{Irrelevance}                      & \multirow{2}{*}{Relevance}                      \\
                      &                                                                             &                                    &                         &                           &                           &                                                                              &                         &                           &                           &                                                                              &                                                   &                                                 \\ \hline
\rowcolor{peacockblue!70} \textbf{1}            & \textbf{87.31}                                                              & {\ul \textbf{xLAM-8x22b-r (FC)}}   & \textbf{72.79}          & \textbf{86.37}            & \textbf{87.13}            & \textbf{84.75}                                                               & \textbf{98.57}          & \textbf{94.00}            & \textbf{92.00}            & \textbf{85.00}                                                               & \textbf{74.96}                                    & \textbf{97.56}                                  \\
2                     & 85.79                                                                       & GPT-4-0125-Preview (Prompt)        & 78.82                   & 88.44                     & 91.00                     & 83.75                                                                        & 99.00                   & 96.00                     & 82.00                     & 80.00                                                                        & 61.35                                             & 97.56                                           \\
3                     & 85.00                                                                       & GPT-4-1106-Preview (Prompt)        & 78.75                   & 89.12                     & 94.12                     & 83.25                                                                        & 99.00                   & 96.00                     & 82.00                     & 72.50                                                                        & 64.98                                             & 90.24                                           \\
4                     & 84.74                                                                       & GPT-4-0613 (Prompt)                & 78.76                   & 85.46                     & 91.75                     & 82.67                                                                        & 98.29                   & 96.00                     & 86.00                     & 70.00                                                                        & 75.57                                             & 82.93                                           \\
5                     & 83.89                                                                       & GPT-4-turbo-20240409 (Prompt)      & 80.47                   & 88.81                     & 88.12                     & 84.25                                                                        & 99.00                   & 96.00                     & 80.00                     & 77.50                                                                        & 61.82                                             & 82.93                                           \\ \rowcolor{peacockblue!55}
\textbf{6}            & \textbf{83.38}                                                              & {\ul \textbf{xLAM-8x7b-r (FC)}}    & \textbf{73.12}          & \textbf{86.09}            & \textbf{71.00}            & \textbf{82.50}                                                               & \textbf{92.57}          & \textbf{96.00}            & \textbf{90.00}            & \textbf{77.50}                                                               & \textbf{72.35}                                    & \textbf{92.68}                                  \\
7                     & 83.35                                                                       & GPT-4o-mini-20240718 (Prompt)    & 75.88                   & 81.64                     & 85.12                     & 79.42                                                                        & 98.29                   & 94.00                     & 82.00                     & 77.50                                                                        & 79.20                                             & 80.49                                           \\
8                     & 83.13                                                                       & GPT-4o-2024-05-13 (Prompt)         & 76.18                   & 86.01                     & 92.12                     & 81.00                                                                        & 98.00                   & 94.00                     & 76.00                     & 72.50                                                                        & 77.44                                             & 78.05                                           \\
9                     & 82.55                                                                       & Functionary-Medium-v3.1 (FC)       & 74.34                   & 87.59                     & 81.62                     & 80.67                                                                        & 98.29                   & 94.00                     & 90.00                     & 75.00                                                                        & 73.23                                             & 70.73                                           \\
10                    & 81.78                                                                       & GPT-4-1106-Preview (FC)            & 69.32                   & 84.19                     & 86.38                     & 71.92                                                                        & 95.43                   & 94.00                     & 86.00                     & 75.00                                                                        & 72.70                                             & 82.93                                           \\
11                    & 81.59                                                                       & Llama3-70B-Instruct (Prompt)      & 72.87                   & 85.91                     & 84.00                     & 77.83                                                                        & 94.14                   & 94.00                     & 84.00                     & 80.00                                                                        & 50.47                                             & 92.68                                           \\
12                    & 80.88                                                                       & Claude-3-Opus (Prompt)             & 76.65                   & 87.47                     & 78.38                     & 75.17                                                                        & 98.57                   & 94.00                     & 82.00                     & 75.00                                                                        & 56.15                                             & 85.37                                           \\
13                    & 80.87                                                                       & GPT-4-0125-Preview (FC)            & 68.76                   & 84.95                     & 80.38                     & 74.00                                                                        & 84.21                   & 94.00                     & 88.00                     & 75.00                                                                        & 74.03                                             & 85.37                                           \\ \rowcolor{peacockblue!45}
\textbf{14}           & \textbf{80.33}                                                              & {\ul \textbf{xLAM-7b-r (FC)}}      & \textbf{69.85}          & \textbf{84.00}            & \textbf{63.00}            & \textbf{79.17}                                                               & \textbf{75.71}          & \textbf{94.00}            & \textbf{92.00}            & \textbf{80.00}                                                               & \textbf{72.88}                                    & \textbf{92.68}                                  \\
15                    & 80.23                                                                       & Nemotron-340b-inst (Prompt)    & 68.51                   & 80.38                     & 78.62                     & 79.17                                                                        & 86.00                   & 90.00                     & 80.00                     & 77.50                                                                        & 84.10                                             & 78.05                                           \\
16                    & 80.21                                                                       & Functionary-Small-v3.1 (FC)        & 72.70                   & 83.31                     & 85.62                     & 72.92                                                                        & 87.79                   & 90.00                     & 86.00                     & 70.00                                                                        & 68.36                                             & 85.37                                           \\ \rowcolor{peacockblue!30}
\textbf{17}           & \textbf{80.18}                                                              & {\ul \textbf{xLAM-7b-fc-r (FC)}}   & \textbf{70.52}          & \textbf{78.22}            & \textbf{73.88}            & \textbf{68.50}                                                               & \textbf{95.21}          & \textbf{90.00}            & \textbf{88.00}            & \textbf{77.50}                                                               & \textbf{79.54}                                    & \textbf{80.49}                                  \\
18                    & 79.66                                                                       & mistral-large-2407 (FC Any)        & 81.01                   & 87.42                     & 90.50                     & 83.50                                                                        & 98.29                   & 92.00                     & 86.00                     & 77.50                                                                        & 0.34                                              & 100.00                                          \\
19                    & 79.55                                                                       & GPT-4o-2024-05-13 (FC)             & 70.40                   & 82.33                     & 89.00                     & 76.08                                                                        & 88.93                   & 84.00                     & 88.00                     & 72.50                                                                        & 73.50                                             & 70.73                                           \\
20                    & 79.25                                                                       & GPT-4o-mini-2024-07-18 (FC)        & 67.83                   & 80.16                     & 85.38                     & 77.17                                                                        & 83.21                   & 92.00                     & 82.00                     & 70.00                                                                        & 71.83                                             & 82.93                                           \\
21                    & 79.14                                                                       & Open-Mixtral-8x22b (Prompt)        & 73.47                   & 76.14                     & 79.12                     & 73.67                                                                        & 91.86                   & 96.00                     & 84.00                     & 75.00                                                                        & 71.42                                             & 70.73                                           \\
22                    & 79.10                                                                       & Gorilla-OpenFunctions-v2 (FC)      & 70.81                   & 79.47                     & 75.75                     & 66.67                                                                        & 95.86                   & 96.00                     & 78.00                     & 70.00                                                                        & 73.13                                             & 85.37                                           \\
23                    & 79.09                                                                       & GPT-4-turbo-2024-04-09 (FC)        & 64.21                   & 82.72                     & 82.50                     & 75.75                                                                        & 88.71                   & 88.00                     & 86.00                     & 72.50                                                                        & 79.79                                             & 70.73                                           \\
24                    & 78.96                                                                       & Functionary-Small-v3.2 (FC)        & 69.50                   & 81.50                     & 80.12                     & 73.50                                                                        & 90.64                   & 88.00                     & 86.00                     & 67.50                                                                        & 72.32                                             & 80.49                                           \\
25                    & 78.87                                                                       & GPT-4o-2024-08-06 (FC)             & 70.71                   & 80.97                     & 83.25                     & 75.58                                                                        & 85.36                   & 90.00                     & 84.00                     & 72.50                                                                        & 82.91                                             & 63.41                                           \\
26                    & 78.78                                                                       & mistral-large-2407 (FC Auto)       & 68.28                   & 86.44                     & 90.25                     & 83.50                                                                        & 76.86                   & 92.00                     & 86.00                     & 77.50                                                                        & 48.93                                             & 78.05                                           \\
27                    & 77.92                                                                       & Claude-3-Sonnet (Prompt)           & 71.80                   & 85.26                     & 82.75                     & 73.92                                                                        & 96.14                   & 90.00                     & 84.00                     & 77.50                                                                        & 30.01                                             & 87.80                                           \\
28                    & 77.45                                                                       & FireFunction-v2 (FC)               & 74.11                   & 81.49                     & 73.62                     & 67.58                                                                        & 94.43                   & 88.00                     & 82.00                     & 72.50                                                                        & 52.94                                             & 87.80                                           \\
29                    & 76.63                                                                       & Granite-20b (FC)   & 65.27                   & 73.05                     & 60.75                     & 67.83                                                                        & 85.36                   & 90.00                     & 84.00                     & 72.50                                                                        & 72.43                                             & 95.12                                           \\
30                    & 76.31                                                                       & Mistral-Nemo-2407 (Prompt)         & 72.89                   & 81.37                     & 81.50                     & 73.75                                                                        & 92.50                   & 94.00                     & 86.00                     & 80.00                                                                        & 13.25                                             & 87.80                                           \\
31                    & 76.29                                                                       & Claude-3.5-Sonnet (Prompt)         & 76.98                   & 80.27                     & 72.62                     & 65.33                                                                        & 98.50                   & 92.00                     & 70.00                     & 72.50                                                                        & 83.46                                             & 51.22                                           \\ \rowcolor{peacockblue!20}
\textbf{32}           & \textbf{75.43}                                                              & {\ul \textbf{xLAM-1b-fc-r (FC)}}   & \textbf{64.63}          & \textbf{72.33}            & \textbf{64.50}            & \textbf{61.42}                                                               & \textbf{80.21}          & \textbf{92.00}            & \textbf{86.00}            & \textbf{75.00}                                                               & \textbf{60.65}                                    & \textbf{97.56}                                  \\
33                    & 75.41                                                                       & GPT-3.5-Turbo (FC)            & 69.79                   & 83.58                     & 71.88                     & 68.83                                                                        & 95.14                   & 88.00                     & 86.00                     & 57.50                                                                        & 35.83                                             & 97.56                                           \\
34                    & 74.97                                                                       & Mistral-Nemo-2407 (FC Auto)        & 64.57                   & 79.99                     & 80.25                     & 74.00                                                                        & 91.36                   & 86.00                     & 86.00                     & 62.50                                                                        & 59.14                                             & 65.85                                           \\
35                    & 74.78                                                                       & Hermes-2-Pro-Llama3-70B (FC)      & 66.29                   & 73.49                     & 70.25                     & 78.33                                                                        & 80.64                   & 88.00                     & 84.00                     & 72.50                                                                        & 53.80                                             & 80.49                                           \\
36                    & 74.75                                                                       & Gemini-1.5-Pro-0514 (FC)   & 56.15                   & 78.89                     & 82.38                     & 65.50                                                                        & 75.71                   & 88.00                     & 84.00                     & 75.00                                                                        & 83.31                                             & 58.54                                           \\
37                    & 74.57                                                                       & Claude-2.1 (Prompt)                & 68.21                   & 78.08                     & 74.12                     & 66.17                                                                        & 94.64                   & 88.00                     & 64.00                     & 62.50                                                                        & 74.36                                             & 75.61                                           \\
38                    & 74.56                                                                       & Gemini-1.5-Pro-0409 (FC)   & 55.08                   & 79.43                     & 83.12                     & 64.75                                                                        & 76.00                   & 88.00                     & 80.00                     & 72.50                                                                        & 83.27                                             & 63.41                                           \\
39                    & 74.12                                                                       & GPT-4o-2024-08-06 (Prompt)         & 65.76                   & 76.86                     & 72.12                     & 71.67                                                                        & 70.57                   & 88.00                     & 78.00                     & 75.00                                                                        & 89.56                                             & 53.66                                           \\
40                    & 74.11                                                                       & Command-R-Plus (Prompt)            & 68.14                   & 78.13                     & 77.50                     & 62.17                                                                        & 91.29                   & 86.00                     & 78.00                     & 55.00                                                                        & 69.31                                             & 75.61                                           \\
41                    & 73.12                                                                       & Mistral-Nemo-2407 (FC Any)    & 67.98                   & 82.46                     & 77.38                     & 76.08                                                                        & 92.07                   & 86.00                     & 86.00                     & 62.50                                                                        & 0.72                                              & 100.00                                          \\
42                    & 72.19                                                                       & Mistral-Medium-2312 (Prompt)       & 63.77                   & 80.22                     & 69.12                     & 59.25                                                                        & 93.43                   & 88.00                     & 70.00                     & 57.50                                                                        & 84.54                                             & 56.10                                           \\
43                    & 72.04                                                                       & Command-R-Plus (FC) (Original)     & 64.25                   & 72.45                     & 66.25                     & 62.33                                                                        & 89.14                   & 86.00                     & 82.00                     & 52.50                                                                        & 52.75                                             & 92.68                                           \\
44                    & 70.75                                                                       & Gemini-1.5-Flash-0514 (FC) & 65.80                   & 83.26                     & 63.87                     & 63.50                                                                        & 57.93                   & 86.00                     & 74.00                     & 75.00                                                                        & 74.69                                             & 63.41                                           \\
45                    & 69.55                                                                       & DBRX-Instruct (Prompt)             & 69.97                   & 80.35                     & 66.88                     & 51.50                                                                        & 90.50                   & 86.00                     & 60.00                     & 62.50                                                                        & 44.86                                             & 82.93                                           \\
46                    & 68.88                                                                       & Claude-3.5-Sonnet (FC)    & 73.95                   & 82.09                     & 65.38                     & 62.75                                                                        & 95.36                   & 86.00                     & 44.00                     & 40.00                                                                        & 75.91                                             & 63.41                                           \\
47                    & 66.19                                                                       & GPT-3.5-Turbo (Prompting)     & 59.01                   & 67.74                     & 65.25                     & 48.58                                                                        & 44.50                   & 86.00                     & 78.00                     & 55.00                                                                        & 69.97                                             & 87.80                                           \\
48                    & 66.18                                                                       & Hermes-2-Pro-Llama3-8B (FC)       & 62.32                   & 74.96                     & 61.62                     & 57.83                                                                        & 68.71                   & 90.00                     & 80.00                     & 57.50                                                                        & 55.16                                             & 53.66                                           \\
49                    & 65.44                                                                       & Hermes-2-Pro-Mistral-7B (FC)       & 60.98                   & 71.49                     & 60.38                     & 50.42                                                                        & 60.50                   & 90.00                     & 84.00                     & 62.50                                                                        & 38.55                                             & 75.61                                           \\
50                    & 64.83                                                                       & Hermes-2-Theta-Llama3-8B (FC)     & 58.53                   & 67.82                     & 59.62                     & 58.33                                                                        & 69.14                   & 88.00                     & 78.00                     & 55.00                                                                        & 62.66                                             & 51.22                                           \\
51                    & 62.70                                                                       & Llama3-8B-Instruct (Prompt)  & 58.53                   & 70.26                     & 53.50                     & 53.25                                                                        & 84.50                   & 88.00                     & 68.00                     & 50.00                                                                        & 22.88                                             & 78.05                                           \\
52                    & 61.89                                                                       & Claude-3-Opus (FC)                 & 69.41                   & 79.95                     & 39.38                     & 27.92                                                                        & 84.64                   & 86.00                     & 52.00                     & 30.00                                                                        & 76.40                                             & 73.17                                           \\
53                    & 60.82                                                                       & Open-Mixtral-8x7b (Prompt)         & 61.49                   & 70.70                     & 47.12                     & 36.83                                                                        & 71.86                   & 74.00                     & 56.00                     & 52.50                                                                        & 71.84                                             & 65.85                                           \\
54                    & 60.34                                                                       & Claude-3-Haiku (Prompt)   & 74.64                   & 84.49                     & 51.88                     & 45.17                                                                        & 89.43                   & 94.00                     & 32.00                     & 27.50                                                                        & 18.90                                             & 85.37                                           \\
55                    & 58.89                                                                       & Open-Mixtral-8x22b (FC Any)        & 73.23                   & 85.42                     & 10.75                     & 63.08                                                                        & 92.57                   & 92.00                     & 24.00                     & 47.50                                                                        & 0.34                                              & 100.00                                          \\ \hline
\end{tabular}
}
\vspace{1mm}
\caption{\small Performance comparison on BFCL-v2 leaderboard (cutoff date 09/03/2024). The rank is based on the overall accuracy, which is a weighted average of different evaluation categories. ``FC" stands for function-calling mode in contrast to using a customized ``prompt" to extract the function calls. See \cite{bfcl} for details.}
\label{table:comparison}
\end{table}

Table \ref{table:comparison} presents the experimental results on the BFCL v2 benchmark (cutoff date 09/03/2024), which shows the exceptional performance of our xLAM model series in function-calling tasks. Notably, xLAM models secure four out of the top twenty positions, demonstrating the effectiveness of our data pipeline and training methodology across various model sizes.

Our flagship model, xLAM-8x22b-r, achieves the highest overall accuracy of 87.31\%, surpassing all other models in the benchmark. This result validates the effectiveness of our data processing and model training pipeline in improving models' function-calling ability. Following closely, xLAM-8x7b-r ranks 6th, outperforming most prominent models including GPT-4o-mini and Claude-3.

The performance of our models demonstrates clear scaling with model size, a trend exemplified by xLAM-7b-r, which ranks 14th with an accuracy of 80.33\%. This model outperforms several larger and more resource-intensive alternatives, including multiple GPT-4 and GPT-4o versions, highlighting the potential of small models in the agent area.

Perhaps most remarkably, our smallest model, xLAM-1b-fc-r, achieves a 32nd place ranking with an accuracy of 75.43\%, surpassing much larger models like Claude-3-Opus (FC) and GPT-3.5-Turbo. This performance underscores the power of our data synthesis framework in producing high-quality, diverse datasets that enhance function-calling effectiveness even for smaller language models.

It is also worth noting that the BFCL v2 benchmark \cite{bfcl} includes a live dataset released after our model training date. These fresh data are collected from real-world user queries that were entirely unseen by our models. Nevertheless, our models exhibit strong generalization capabilities in handling these real-world use cases. 
The consistently strong performance across our model series, ranging from 8x22 billion to 1 billion parameters, demonstrates the scalability and versatility of our approach. This scalability is particularly noteworthy, as it enables strong results from compact models suitable for resource-constrained environments to large-scale models for more demanding applications. Furthermore, the ability of our smaller models to compete with much larger alternatives suggests significant potential for efficient deployment in various real-world scenarios.

\subsection{Ablation Study}
\begin{wrapfigure}{r}{0.55\textwidth}
  \begin{center}
\includegraphics[width=0.99\linewidth]{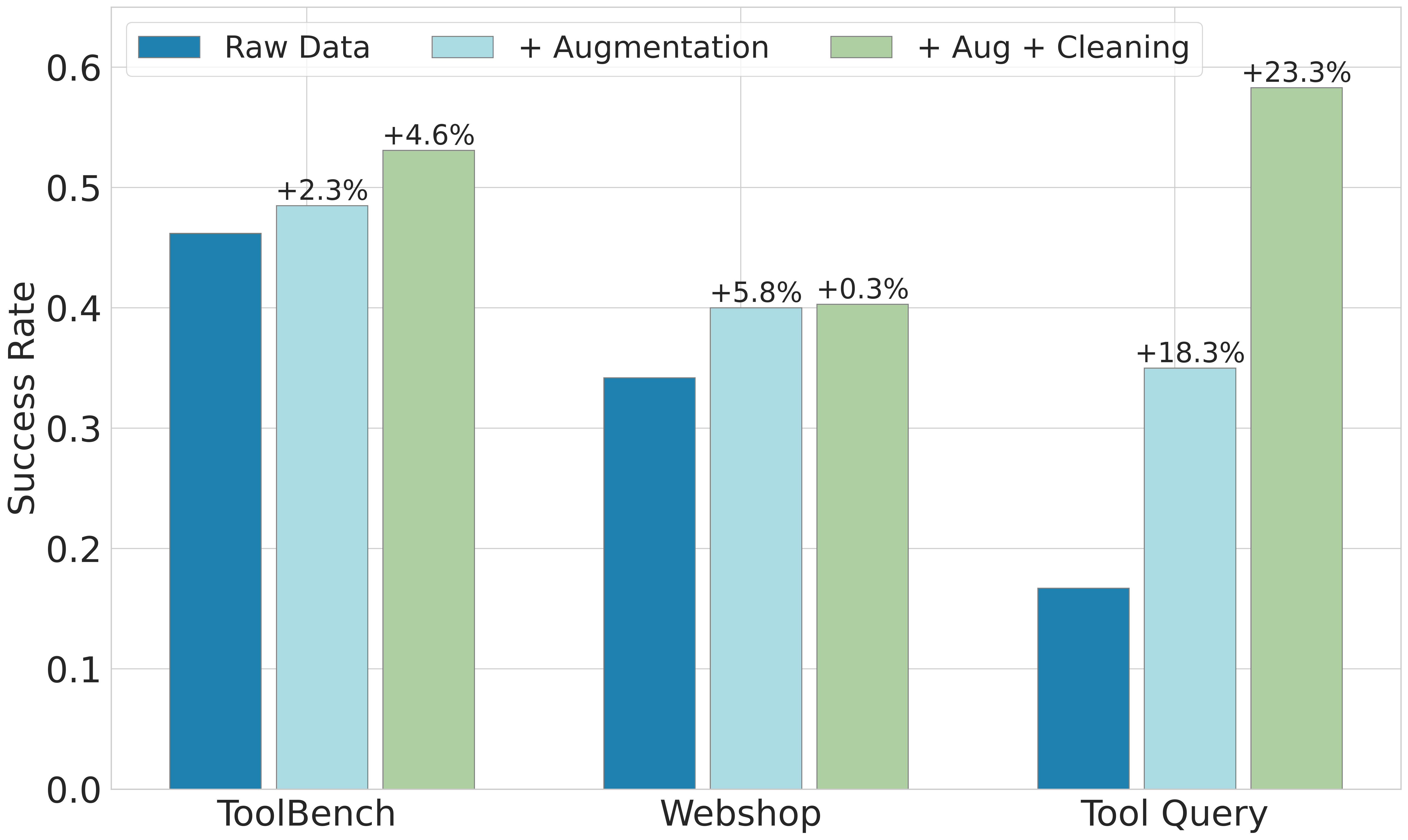}
  \end{center}
    \caption{\small Ablation study for data augmentation and data quality verification (cleaning).}
  \label{fig:ablation}
  \vspace{-3mm}
\end{wrapfigure}

We conducted an ablation study on the 7B models to measure the impact of various steps in our data pipeline.
Three datasets were prepared for this analysis: raw data, augmented data, and augmented + cleaned data. The raw data represents the dataset before data unification, while the other two datasets are post-unification.
Figure \ref{fig:ablation} presents the evaluation results of models trained on these three datasets. The metrics used for this evaluation are G1\_instruction from ToolBench and success\_rate from both Webshop and ToolQuery. The results indicate that augmented data consistently outperforms raw data across all metrics, with improvements of 2.3\% on ToolBench, 5.8\% on Webshop, and 18.3\% on ToolQuery. Furthermore, the addition of data cleaning leads to a substantial performance increase on ToolQuery, with a further improvement of 23.4\%. The results highlight the effectiveness of data augmentation and cleaning processes in the data pipeline.

\section{Conclusion}

This paper introduces xLAM series, a set of large action models for autonomous AI agents. Our models, ranging from 1B to 8x22B parameters, were trained with a scalable and flexible data pipeline that unifies, augments, and synthesizes diverse datasets. 
Our evaluations show that xLAM models consistently perform exceptionally across various benchmarks. 
The insights we learned from training these models highlight the importance of rigorous data processing and the potential of data synthesis in developing capable AI agents.
By releasing the xLAM series to the public, we aim to democratize access to high-performance models for agent tasks, thereby accelerating progress in the field.

\bibliographystyle{unsrt}
\bibliography{main}

\clearpage
\appendix
\section{Appendix}

\begin{figure}[ht]
\begin{lstlisting}[language=json,basicstyle=\scriptsize\ttfamily, backgroundcolor=\color{lightgold!50}]
{
    "unique_trajectory_id": "id",
    "task_instruction": "...",
    "few_shot_examples": [],
    "query": "The task or the question that the user provides.",
    "tools": [
        {
            "name": "api_name1",
            "description": "description of this api",
            "parameters": {
                "param1": {
                    "type": "string",
                    "description": "",
                },
            }
        },
    ],
    "steps": [
        {
            "thought": "thinking and/or planning process",
            "tool_calls": [
                {
                    "name": "api_name1",
                    "arguments": {
                        "argument1": "xxx.",
                        "argument2": "xxx"
                    }
                }
            ],
            "step_id": 1,
            "next_observation": "observations or feedbacks from the environment/APIs after execution function."
            "user_input": "User follow up input at this turn if any."
        },
    ],
}
\end{lstlisting}
\caption{Unified function calling data format.}
\label{fig:unified_format}
\end{figure}

\tcbset{
    mybox/.style={
        colback=lightgold!50,
        colframe=lightgold!50,
        arc=4mm,
        boxrule=0.5mm,
        left=1mm,
        right=1mm,
        top=1mm,
        bottom=1mm,
        boxsep=1mm,
        coltitle=black,
        sharp corners,
    }
}
\begin{figure*}[htbp]

\begin{tcolorbox}[mybox,  fonttitle=\bfseries\footnotesize, before upper=\small]
\textbf{Prompt:}
\begin{verbatim}
[BEGIN OF TASK INSTRUCTION]
Based on the previous context and API request history, generate an API 
request or a response as an AI assistant. 
[END OF TASK INSTRUCTION]

[BEGIN OF AVAILABLE TOOLS]
[
    {
        "name": "get_fire_info",
        "description": "Query the latest wildfire information",
        "parameters": {
            "location": {
                "type": "string",
                "description": "Location of the wildfire.",
                "required": true,
            },
            "radius": {
                "type": "number",
                "description": "The radius (in miles) around the location.",
                "required": false,
            }
        },
    },...
]
[END OF AVAILABLE TOOLS]

[BEGIN OF FORMAT INSTRUCTION]
Your output should be in the JSON format, which specifies a list of
function calls. The example format is as follows. Please make sure the 
parameter type is correct. If no function call is needed, please make 
tool_calls an empty list "[]".
"""
{"thought": "the thought process, or an empty string", "tool_calls": 
[{"name": "api_name1", "arguments": {"argument1": "value1", "argument2":
"value2"}}]}
"""
[END OF FORMAT INSTRUCTION]

[BEGIN OF QUERY]
Can you give me the latest information on the wildfires occurring in California?
[END OF QUERY]

[BEGIN OF HISTORY STEPS]
[
    {
        "thought": "Sure, what is the radius (in miles) around the 
        location of the wildfire?",
        "tool_calls": [],
        "step_id": 1,
        "next_observation": "",
        "user_input": "User: Let me think... 50 miles."
    },
]
[END OF HISTORY STEPS]
\end{verbatim}

\textbf{Output:}
\begin{verbatim}
{"thought": "", "tool_calls": [{"name": "get_fire_info", 
"arguments": {"location": "California", "radius": 50}}]}
\end{verbatim}

\end{tcolorbox}
\caption{\small Example prompt and output for function-calling using xLAM.}
\label{fig:training_template}
\end{figure*}

\end{document}